# CNN-based search model underestimates attention guidance by simple visual features

Endel Põder

Recently, Zhang et al. (2018) proposed an interesting model of attention guidance that uses visual features learnt by convolutional neural networks for object recognition. I adapted this model for search experiments with accuracy as the measure of performance. Simulation of our previously published feature and conjunction search experiments revealed that CNN-based search model considerably underestimates human attention guidance by simple visual features. A simple explanation is that the model has no bottom-up guidance of attention. Another view might be that standard CNNs do not learn features required for human-like attention guidance.

## Introduction

Visual search is a usual part of our everyday activity, and a widely used method in vision research. Dependent on the target and visual environment, search efficiency can be very different.

Search is efficient when pre-attentively available visual features direct attention to the location of the target object, and inefficient when there are no pre-attentive features that could easily discriminate the target from its surroundings (e.g., Wolfe et al., 1989; Wolfe & Horowitz, 2017).

Traditionally, it has been assumed that attention is guided by a relatively small number of simple features like brightness, color, size, orientation, movement (e.g., Wolfe & Horowitz, 2004).

Recently, Zhang et al. (2018) proposed an interesting model of attention guidance that uses complex visual features learnt by convolutional neural networks for object recognition tasks. These features exhibit some invariance relative to simple transformations like position, size, and rotation. To be useful for guidance of spatial attention, still certain extent of position selectivity is required. Zhang et al. (2018) found that the last convolutional layer of VGG16 neural net has required properties. It has low spatial resolution (14x14), and a feature vector composed of 512 features. When modeling visual search, Zhang et al. (2018) presented an image with the target object in its center at first, and saved feature vector corresponding to the target from the central position of layer 31 in VGG16. Then search display with the target among distractors was presented and dot product of the saved target vector with feature vector from each of 14x14 spatial position of layer 30 was calculated. The resultant attention map was used to select directions for eye movements. This model accounted well for human eye movements and reaction times when searching for a target among natural objects, or in cartoon pictures.

However, Zhang et al. (2018) did not test their model with simple classic search tasks (e.g., Treisman & Gelade, 1980). In some experiments, they even tried to remove simple features like color, brightness, orientation that might direct attention in natural conditions. A general model of attention guidance should accurately predict data from simple classic search experiments as well.

The original Zhang et al. (2018) model was built to predict eye movements. However, the same attention map can be naturally used to predict proportion correct in experiments with brief presentation of stimuli. We just need to replace the target localization rules with a yes-no detection one.

In this study, I apply Zhang et al. (2018) model to simple feature and conjunction search stimuli from our resent study (Põder & Kosilo, 2019), and compare the simulation results with the results from human observers. Relatively rich data set from Põder and Kosilo (2019) experiments allows to judge model's behavior in several respects.

## Methods

### Stimuli

Examples of stimuli are given in Figure 1. Stimuli from Põder and Kosilo (2019) were designed considering properties of human peripheral vision. There are aspects that may be unnecessary or problematic when experimenting with CNN. Thus, I removed the ellipticity of display and eccentricity-dependency of inter-item distances. However, the overall configuration with concentric rings of stimuli was retained.

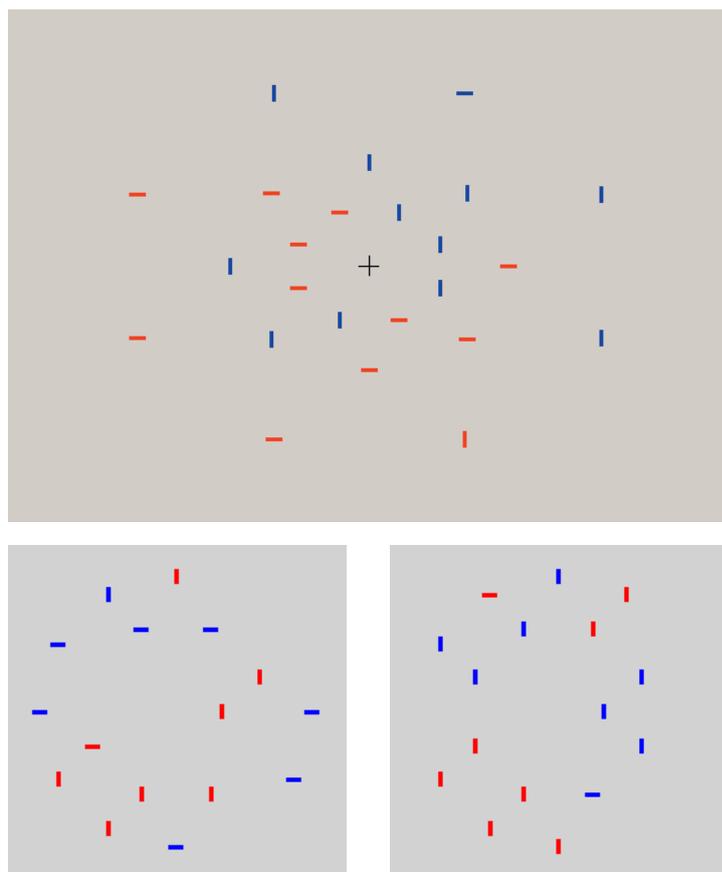

Figure 1. Examples of stimuli. A display for conjunction search from Põder and Kosilo (2019) (top), and stimuli used in the present study (bottom): conjunction search (left), feature search (right). Both stimuli are with target present, set size 16, and bar length 10 pixels.

Põder and Kosilo (2019) used two targets in target-present trials to exactly balance feature distributions in conjunction search displays. According to observers' comments and some regularities of response distributions observers still looked for one target they had chosen. Similarly, our simulation model searched for one fixed target from the two options available.

The main parameters – set size and length of line segments, were varied over the ranges used in Põder and Kosilo (2019) study, set size from 2 to 24, and bar length from 4 to 17 pixels (bar width was 3 pixels).

## Simulations

For simulations, I used a pretrained VGG16 network available from Matlab Deep Learning toolbox. I used my target and search stimuli as input and registered corresponding activations from layers 30 or 31. At first, an image of the target in the center of display was presented, and the activation vector from the center of the layer 31 was saved. Then, search images were presented, and all activations from layer 30 were saved. The dot product of the target vector with layer 30 activation vectors, in each spatial position, was calculated. The result is 14x14 attention map.

Because Põder and Kosilo (2019) experiments used a brief presentation of stimuli and proportion correct as measure of performance, simulation of eye movements is not required. Instead, model should use activations in the attention map to judge whether target was present or not. I used max rule that closely approximates the ideal decision rule. When maximum value of attention map exceeds the criterion, then model's response is "target present", otherwise – "target absent".

In Matlab, it is natural to operate with arrays of stimuli rather than run trial-by-trial simulations. I presented arrays of stimuli to the neural network and saved the corresponding distributions of output activations, separately for target present and target absent trials. It could be easy to use means and standard deviations to calculate d prime and transform that into proportion correct. However, these results likely underestimate correct values because the underlying distributions can be far from normal. Therefore, I used histograms of attention map maxima, for target present and target absent trials, and searched for the criterion giving the best performance. However, using the same data to estimate the criterion and respective performance could cause a kind of overfitting. Thus, in the present simulations I used double runs: the first to estimate optimal criterion, and second to calculate proportion correct using that criterion. The graphs are based on 1000 simulated trials per data point.

## Results

The results of simulation, for feature and conjunction search, together with human results from Põder and Kosilo (2019) are given in Figure 2. In several respects, simulated and human results are similar. Both exhibit typical psychometric curves as dependent on bar length. Also, performance is much lower in conjunction search compared to feature search, for both humans and model. However, there are clear differences as well. Overall performance of the model is much worse compared to human observers, and the differences between feature vs conjunction search are less dramatic in the model. While human observers can detect a horizontal bar among vertical ones almost perfectly, independent of the number of bars, once the bar length exceeds twice its width, the model exhibits no such pop-out performance. Even with very long bars, performance remains far from perfect, and depends heavily on set-size.

While the simulation results for one target (red horizontal) are given in Figure 2, similar results have been found for different targets and different distractors. Also, the presence/absence of the complementary target has very little effect – comparable curves can be produced using traditional displays with a single search target.

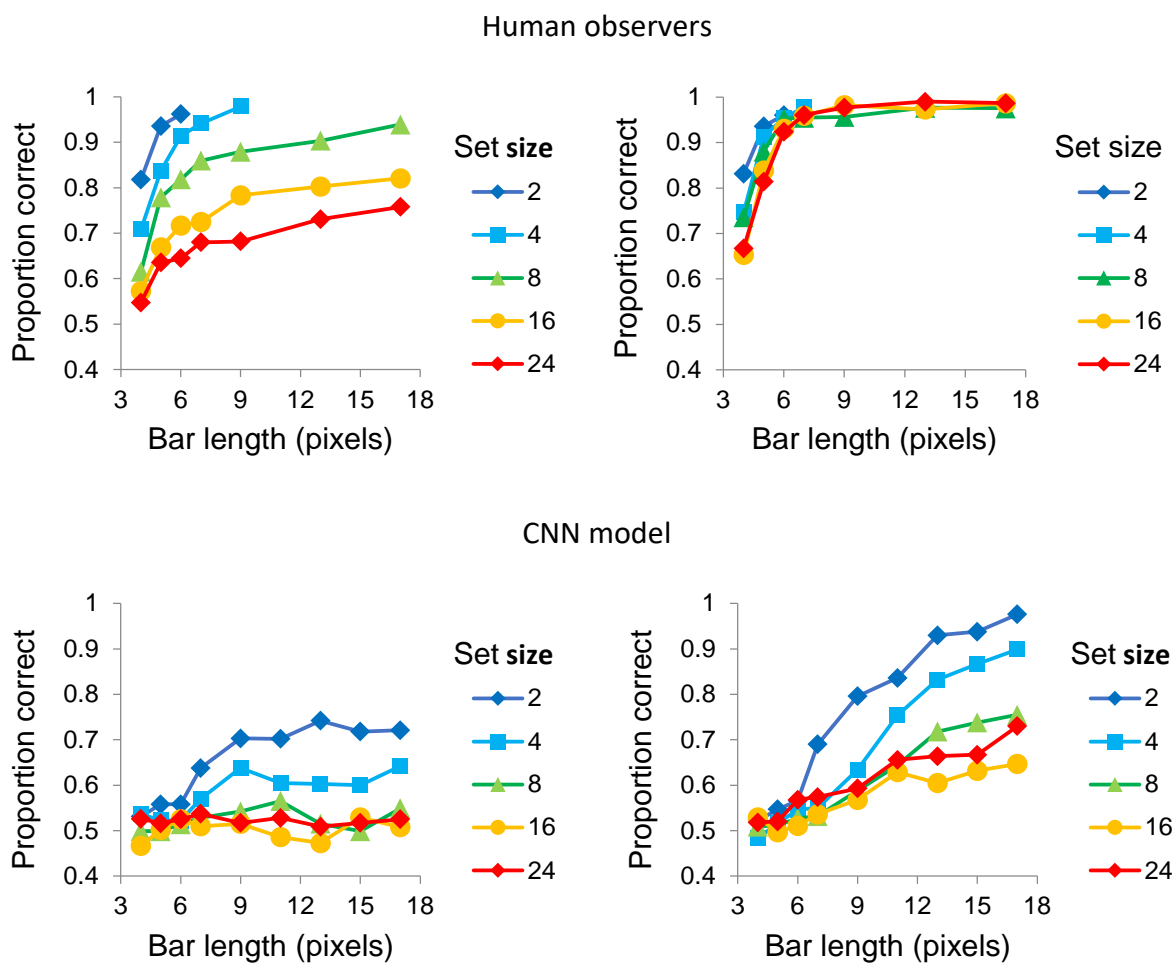

Figure 2. Results of search experiments with human observers (from Põder & Kosilo, 2019), and with CNN based search model. Conjunction search (left) and feature search (right).

## Discussion

Zhang et al (2018) found that complex features available in the last retinotopic layer of VGG16 neural net account well for guidance of eye movements when searching for natural objects as targets. The present study reveals that Zhang et al. (2018) model cannot explain human visual search for simple features, and their conjunctions. Apparently, the features from that VGG16 layer do not contain information required to efficiently direct attention to the target in a simple feature search.

A simple explanation could be that there is no bottom-up guidance in Zhang et al. (2018) model. Classic models of visual search assume that there are two mechanisms of attention guidance – top-down, and bottom-up (Cave & Wolfe, 1990; Wolfe & Horowitz, 2017). The first compares features from an input image with target features, the second calculates visual saliency of input features independent of the search target. Zhang et al. (2018) model apparently incorporates only the first. Because bottom-up guidance played no important role with their stimuli, the model predicted human performance reasonably well. However, with simple stimuli, bottom-up saliency is much more important, but the model ignores that.

Perhaps it is possible to include saliency calculation as an additional component in attention guidance mechanism. However, saliency calculation might be a regular component within the feature extraction network instead (e.g., Li, 2002). In that case, we could blame pretrained VGG16 network for not providing the required set of bottom-up features. Both options may be consistent with my earlier study that revealed a remarkable inefficiency of simple feature search in pretrained CNNs (Põder, 2017, 2019).

Anyway, the Zhang et al. (2018) model incorporates only a part of human attention guidance.

## Acknowledgements

I thank Jaan Aru and Tarun Khajuria for the idea to test Zhang et al. (2018) model with simple feature and conjunction stimuli.